\documentclass{article}

\usepackage{microtype}
\usepackage{graphicx}
\usepackage{subfigure}
\usepackage{booktabs} 

\usepackage{hyperref}

\usepackage[preprint]{icml2024}

\usepackage{amsmath}
\usepackage{amssymb}
\usepackage{mathtools}
\usepackage{amsthm}

\usepackage[capitalize]{cleveref}
\crefname{section}{Sec.}{Sec.}

\theoremstyle{plain}

\theoremstyle{definition}

\theoremstyle{remark}

\newcommand{\squish}[1]{{#1\parfillskip=0pt\par}}

\usepackage[textsize=tiny]{todonotes}

\icmltitlerunning{Classical Statistical (In-Sample) Intuitions Don't Generalize Well}

\begin{document}

\twocolumn[
\icmltitle{Classical Statistical (In-Sample) Intuitions Don't \textit{Generalize} Well: A Note on Bias-Variance Tradeoffs, Overfitting and Moving from Fixed to Random Designs}



\begin{icmlauthorlist}
\icmlauthor{Alicia Curth}{yyy}

\end{icmlauthorlist}

\icmlaffiliation{yyy}{ University of Cambridge}

\icmlcorrespondingauthor{Alicia Curth}{amc253@cam.ac.uk}

\icmlkeywords{Bias-variance tradeoff}

\vskip 0.3in
]



\printAffiliationsAndNotice{}  

\begin{abstract}\vspace{-.1cm}
The sudden appearance of modern machine learning (ML) phenomena like double descent and benign overfitting may leave many classically trained statisticians feeling uneasy -- these phenomena \textit{appear to} go against the very core of statistical intuitions conveyed in any introductory class on learning from data. The historical lack of earlier observation of such phenomena is usually attributed to today's reliance on more complex ML methods, overparameterization, interpolation and/or higher data dimensionality. In this note, we show that there is another reason why we observe behaviors today that appear at odds with intuitions taught in classical statistics textbooks, which is much simpler to understand yet rarely discussed explicitly. In particular, many intuitions originate in \textit{fixed design} settings, in which in-sample prediction error (under resampling of noisy outcomes) is of interest, while modern ML evaluates its predictions in terms of \textit{generalization error}, i.e. out-of-sample prediction error in \textit{random designs}. Here, we highlight that this simple move from fixed to random designs has (perhaps surprisingly) far-reaching consequences on textbook intuitions relating to the bias-variance tradeoff, and comment on the resulting (im)possibility of observing double descent and benign overfitting in fixed versus random designs. 
\end{abstract}

\vspace{-.9cm}
\section{Introduction}
The strikingly good performance of highly overparametrized machine learning (ML) models trained to zero loss, ubiquitously observed in recent years \citep{neyshabur2014search, bartlett2020benign, zhang2021understanding, belkin2021fit}, appears to contradict all classical statistical intuitions about overfitting -- and as such, probably leaves many classically trained statisticians confused and somewhat uneasy. This definitely applies to myself, and has left me puzzled, questioning a lot of what we are taught in graduate statistics classes.  Should we no longer be concerned about overfitting and bias-variance tradeoffs? If modern ML methods defy the intuitions built in decades of statistics research, is there something that has changed recently? Or is modern ML simply a magic bullet we had been missing all along? 

This note therefore aims to better understand the \textit{sources} of the discrepancies between classical statistical intuitions surrounding the bias-variance tradeoff and modern ML phenomena like double descent \cite{belkin2019reconciling} and benign overfitting \cite{bartlett2020benign}. The historical lack of earlier observation of such phenomena is usually attributed to today's reliance on more complex ML methods, overparameterization, interpolation and/or higher data dimensionality \cite{belkin2021fit}. Here, we will explore another reason why we observe behaviors today that appear at odds with intuitions taught in classical statistics textbooks, which is much simpler to understand yet rarely discussed explicitly. We highlight that statistics historically focussed on \textit{fixed design} settings \cite{rosset2019fixed}, where in-sample prediction error is of interest, while modern ML evaluates its predictions in terms of \textit{generalization error}, i.e. out-of-sample prediction error -- and this seemingly small change has surprisingly far-reaching effects on textbook intuitions. 

\squish{\textbf{Outlook.} \cref{sec:setup} introduces fixed and random design setups. In \cref{sec:bias-var}, we  revisit the bias-variance tradeoff. Using a simple k-Nearest Neighbor estimator on low-dimensional data, we show that the classical bias-variance tradeoff intuition (\textit{``Variance increases with model complexity, Bias decreases with model complexity"}) does not necessarily hold when considering out-of-sample prediction error: there can exist regimes where \textit{both} bias and variance decrease when complexity is \textit{decreased}. That is, we show that classical intuitions relating bias, variance and model complexity break \textit{already in the absence} of modern ML methods, overparameterization, interpolation and high-dimensional data, highlighting that they cannot be solely responsible for the emergence of surprising statistical phenomena. In \cref{sec:dd} we then comment on the recent appearance of double descent, and show that one reason for the historical absence of double descent shapes in the literature may be that they \textit{cannot} appear in fixed design settings.  In \cref{sec:overfitting}, we comment on benign overfitting, and discuss when and why it is possible.}

\section{Problem setup: Fixed vs Random designs}\label{sec:setup}
It appears that much of the statistics literature has historically focussed on so-called fixed design settings \citep{rosset2019fixed}, where \textit{in-sample} prediction error (which assumes that test-time inputs will be the same as training inputs but noisy labels are \textit{re-sampled}) is used to measure test-time model performance. The modern ML literature, on the other hand, is almost exclusively interested in \textit{generalization} to new inputs \citep{goodfellow2016deep, murphy2022probabilistic} -- i.e. out-of-sample prediction error, where \textit{both} test inputs and test labels are newly sampled. 

Formally, as in e.g. \citet{rosset2019fixed}, assume we observe a sample of labeled training observations $\{(x_i, y_i)\}^n_{i=1}$, consisting of pairs of inputs $x \in \mathcal{X } \subset \mathbb{R}^d$ and outcomes $y \in \mathbb{R}$, which are jointly sampled i.i.d. from some distribution $P$. Outcomes are related to inputs as $y = f^*(x) + \epsilon$ where $f^*(x) = \mathbb{E}[y|x]$ and we assume that $\epsilon = y - f^*(x)$ is independent of $x$, implying homoskedastic variance $\sigma^2=\text{var}(y|x)$.

{\textbf{Setting A: Fixed design.} In a fixed design setting it is assumed that \textit{the same} inputs $\{x_i\}^n_{i=1}$ as the training inputs\footnote{Classical fixed design settings also assume that the $x_i$ are non-random (e.g. because they were designed). Above, we allow randomness in the $x_i$ and only require train- and test-time realizations of $x_i$ to be the same. \citet{rosset2019fixed} refer to this as the 'Same-x' setting. For the purpose of this note, making a distinction between the two is not necessary.} are encountered at test-time, but with new realizations of the outcomes $\{\tilde{y}_i\}_{i=1}^n$ drawn independently from the conditional law of $y_i|x_i$. If we learn a predictor $\hat{f}(\cdot): \mathcal{X } \rightarrow \mathbb{R}$ from the training data, we thus ultimately want to minimize its in-sample prediction error }
\begin{equation}
    ERR_{is} = \mathbb{E}_{\tilde{y}}\left[\frac{1}{n}\sum^n_{i=1}(\tilde{y}_i - \hat{f}(x_i))^2\right]
\end{equation}

\textbf{Setting B: Random design (generalization).} In a random design setting, we are instead interested in \textit{generalization} of our learned predictor $\hat{f}(\cdot)$ to \textit{new} inputs $x_0$ that have \textit{not} been observed during training, but are also randomly sampled at test-time; i.e. both input and output are newly sampled as $(x_0, y_0) \sim P$.  We thus ultimately want to minimize the out-of-sample prediction (or: generalization) error:

\begin{equation}
    ERR_{oos} = \mathbb{E}_{x_0, y_0}\left[(y_0 - \hat{f}(x_0))^2\right]
\end{equation}

\section{How the move to random design settings affects the bias-variance tradeoff}\label{sec:bias-var}
Below, we now  explore how changing from fixed to random design settings affects our intuitions around the bias-variance tradeoff. Using a simple example with k-nearest neighbor estimators, we highlight that -- perhaps surprisingly -- the classical intuition around the bias-variance tradeoff (\textit{"Variance increases with model complexity, bias decreases with model complexity"}, see below) does not necessarily hold up when we consider the random design setting instead of the classical fixed design. 

\textbf{Preliminaries: k-Nearest Neighbor estimators.} To show this, we examine the bias and variance of very simple estimators -- k-Nearest Neighbor (k-NN) estimators -- to highlight that surprising behaviors of bias and variance are \textit{not actually unique} to more complex modern ML methods. Recall that, for any input $x$, a k-NN estimator issues predictions that are averages of outcomes across the $k$ nearest training examples whose indices are collected in $N^k(x)$, i.e. 
\begin{equation}
   \textstyle    \hat{f}(x) = \frac{1}{k}\sum^n_{i=1} \mathbf{1}\{i \in N^k(x)\} y_i = \sum^n_{i=1} w_{k,i}(x) y_i
\end{equation}
{Here, we sometimes collect the nearest neighbor weights in the $k\times 1$ vector $\mathbf{w}_k(x)=[w_{k,1}(x), \ldots, w_{k,n}(x)]=[\frac{1}{k} \mathbf{1}\{1 \!\in \!N^k(x)\},\ldots,  \frac{1}{k}\mathbf{1}\{n \!\in\! N^k(x)\}]$. Recall also that complexity in k-NN estimators is \textit{inversely} related to $k$: the 1-NN estimator is the most complex estimator in this class while the n-NN estimator is simply the sample mean.  }

\textbf{Expressions for bias and variance.} For some test input $x \in \mathcal{X}$, the bias and variance of a k-NN estimator for given training inputs $\{x_i\}^n_{i=1}$ 
are \citep[Ch. 7.3]{hastie2009elements}
\begin{equation}\label{eq:bias}
    \textstyle    \text{Bias}_{k}(x) = f^*(x) - \sum^n_{i=1} w_{k, i}(x)f^*(x_i)
\end{equation}
\begin{equation}\label{eq:var}
          \text{Var}_k(x) =\text{Var}_k = \frac{\sigma^2}{k}
\end{equation}

\subsection{The classical bias-variance tradeoff intuition: in-sample view} \squish{Statistics textbooks often discuss the bias-variance tradeoff by considering what happens to bias and variance of predictions \textit{for previously observed training inputs} as $k$ increases \citep[Ch. 3.3]{hastie1990GAM}. For the variance term, this is easily read off from \cref{eq:var}: the variance of predictions due to noise in the training labels $y_i$ is always monotonically decreasing in $k$.  When considering how the bias of k-NN estimators {at a training input} $x_j$ is likely to evolve, note that a 1-NN estimator -- which has $N^1(x_j)=\{j\}$ and hence $\mathbf{w}_1(x_j)=\mathbf{e}_j$ (with $\mathbf{e}_j$ the j-th unit vector) -- will always have \textit{no bias}, as $\sum^n_{i=1} w_{1,i}(x_j)f^*(x_i) = f^*(x_j)$. As $k$ increases, this bias is likely to increase because the weighted average $\sum^n_{i=1} w_{k,i}(x_j) f^*(x_i)$ involves more terms with $f^*(\cdot)$ different from  $f^*(x_j)$ \citep[Ch. 3.3]{hastie1990GAM}. }

This is precisely the intuition behind the bias-variance tradeoff as presented in e.g. \citet[Ch. 3.3]{hastie1990GAM} and \citet[Ch. 7.3]{hastie2009elements}: As the complexity of the estimator increases (k decreases), variance is expected to monotonically increase while bias is expected to monotonically decrease\footnote{Note that the term \textit{bias-variance tradeoff} is not to be confused with the term \textit{bias-variance decomposition}. The term bias-variance decomposition \citep[Ch. 7.3]{hastie2009elements} refers to the \textit{fact} that (for any estimator $\hat{f}$) the mean squared error of estimation can always be \textit{decomposed} into a squared bias and a squared variance term, i.e. $\mathbb{E}[(f^*(x) - \hat{f}(x))^2]=Bias^2(x)+Var(x)$. Further, the decomposition of the mean squared prediction error incurs an additional term due to noise in outcome, i.e. $\mathbb{E}[(y- \hat{f}(x))^2]=Bias^2(x)+Var(x) + \sigma^2$.}.

\subsection{New territories: Bias-bias-variance tradeoffs in random design settings}\label{sec:bias-bias-var}

\begin{figure}[t]
    \centering
    \includegraphics[width=0.99\linewidth]{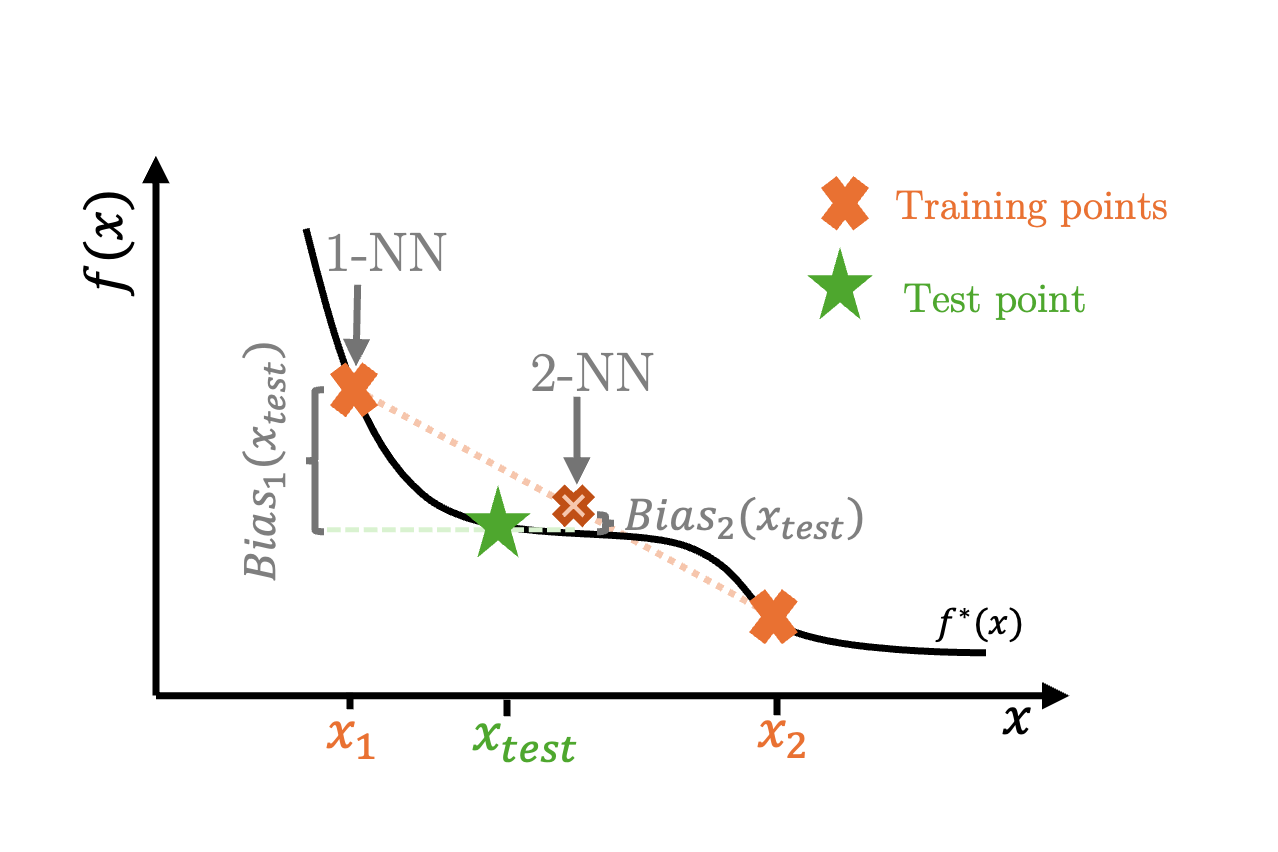}
    \caption{\textbf{Stylized example: The 1-NN estimator does not necessarily have the lowest bias when considering test inputs different from training inputs.}}
    \label{fig:nn-illustration}
\end{figure}

{Next, we highlight that the bias-variance tradeoff intuition does not necessarily hold up even for simple k-NN estimators once we move to the random design setting. This is because bias \textit{no longer monotonically decreases} as complexity increases. Intuitively, this is because -- as there is no training point $x_j$ with exactly the same input value as the new test point $x_0$ -- there generally is \textit{no perfect match with zero bias} among the neighbors: that is, there does not necessarily exist any training example  $x_j$ so that $f^*(x_0)=f^*(x_j)$, which means that the 1-NN estimator does not necessarily have the lowest bias. This intuition is illustrated in \cref{fig:nn-illustration} using a stylized example.}

To make this more formal, we note that, by adding \textit{and} subtracting $f^*\Big(\textstyle \sum^n_{i=1} w_{k, i}(x_0) x_i\Big)$ from \cref{eq:bias}, it is always possible\footnote{Note that this decomposition is useful not only for k-NN estimators but for any estimator issuing predictions that are weighted averages of training outcomes, which includes the broad class of regression smoothers \cite{hastie1990GAM}.} to rewrite the bias term as\footnote{We borrow the idea for this decomposition from the causal inference literature where it appeared in \citet{kellogg2021combining} when comparing the biases of matching- and synthetic control estimators.
}:
\begin{equation*}
\begin{split}
        \text{Bias}_{k}(x_0) = \underbrace{\bigg(f^*(x_0) - f^*\Big(\textstyle \sum^n_{i=1} w_{k, i}(x_0) x_i\Big) \bigg)}_{\text{NeighborMatchingBias}_k(x_0)} + \\ \underbrace{ \bigg(f^*\Big(\textstyle\sum^n_{i=1} w_{k, i}(x_0) x_i\Big) -  \sum^n_{i=1} w_{k, i}(x_0) f^*(x_i)\bigg)}_{\text{AveragingBias}_k(x_0)}
\end{split}
\end{equation*}

This is interesting because the first term captures the bias in prediction arising due to mismatches between the test example and the selected neighbors in \textit{input space}; this term is zero whenever the neighbor weights reconstruct the test input perfectly as $x_0=\textstyle \sum^n_{i=1} w_{k, i}(x_0) x_i$. The second term captures the bias arising in any estimator that predicts using \textit{weighted averages} due to nonlinearity of the true (unknown) prediction function $f^*$: if $f^*$ was linear, then the averaging operation and application of the function would commute (in which case an estimator perfectly reconstructing the inputs will incur zero bias). Note that a 1-NN estimator incurs \textit{no} averaging bias because $\mathbf{w}_1(x)$ has only one nonzero element, but could incur significant neighbor matching bias if $x_0$ is far from its nearest training neighbor $x_{j^*}$. A k-NN estimator with average input $\textstyle \sum^n_{i=1} w_{k, i}(x_0) x_i$ closer to $x_0$ in input space than its nearest neighbor $x_{j^*}$ would incur \textit{less} neighbor matching bias but may incur significant averaging bias depending on the nonlinearity of the underlying $f^*(x_i)$. 

When considering \textit{in-sample} prediction at training point $x_j$, the 1-NN estimator has $\textstyle \sum^n_{i=1} w_{1, i}(x_j) x_i=x_j$; thus both bias terms are exactly zero. For out-of-sample prediction, the $\text{NeighborMatchingBias}_1(x_0)$ term can be substantial depending on the distance of any input to its nearest neighbor in the training data. A k-NN estimator with $k>1$ \textit{can} improve this component of the bias, but will likely worsen the averaging bias (if the underlying $f^*$ is nonlinear). This is precisely the reason why the bias term no longer necessarily behaves monotonically in the random design setting and is illustrated in \cref{fig:nn-illustration}.

\subsection{Empirical investigation: How do bias terms evolve in- and out-of-sample?}
Next, we empirically investigate whether these theoretical predictions indeed hold up: does the bias term behave differently in- and out-of-sample? Here, we use a nonlinear DGP adapted from \citet{friedman1991multivariate}, with 
\begin{equation}\label{eq:marsmult}
f^*(x)=10sin(\pi x_1x_2) + 20 (x_3 - \frac{1}{2})^2 + 10 x_4+ x_5
\end{equation}
and let $y=f^*(x)+\epsilon$, where $\epsilon \sim \mathcal{N}(0, \sigma^2)$ and the $d\!\!=\!\!5$ features $x\in [0, 1]^d$ are sampled independently from $Unif(0,1)$. We use $n=100$ as training data and sample $100$ further out-of-sample test examples.

\begin{figure}[t]
    \centering
    \includegraphics[width=.99\columnwidth]{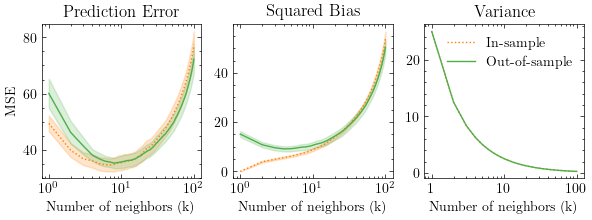}
    \caption{\textbf{The classical bias-variance tradeoff occurs in in-sample prediction error, but not in out-of-sample prediction error -- where decreasing $k$ can decrease \textit{both} bias and variance.} The behavior of Prediction error, Bias and Variance by k for kNN estimators, in-sample (orange) and out-of-sample (green).  Data simulated using $f^*(x)$ from \cref{eq:marsmult} with $\sigma=5$.}
    \label{fig:bias-var}
\end{figure}

\begin{figure}[b]
    \centering
    \includegraphics[width=.99\columnwidth]{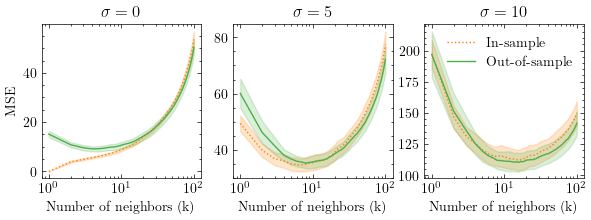}
    \caption{\textbf{Bias alone can cause the U-shape in out-of-sample prediction error (while in-sample the U-shape is caused by the bias-variance tradeoff and thus appears only when $\sigma>0$).} \small The behavior of prediction error by k for kNN estimators, in-sample (orange) and out-of-sample (green) across different levels of noise in outcomes $\sigma$.}
    \label{fig:sigma-effect}
\end{figure}

\paragraph{Non-monotonic behavior of out-of-sample bias.} In \cref{fig:bias-var}, we plot prediction error ($ERR_{is}$ and $ERR_{oos}$), squared bias and variance while simulating data with $\sigma=5$. We observe that the in-sample terms behave as expected (orange): Bias monotonically decreases in complexity (increases in $k$), while variance monotonically increases in complexity (decreases in $k$). This leads to the classical U-shaped tradeoff in in-sample prediction error $ERR_{is}$. While the variance behaves similarly out-of-sample, the bias term shows a strikingly different behavior: there is a U-shape in the bias itself, as the most complex 1-NN estimator indeed does not have the lowest bias. Instead, intermediate values of $k$ incur lowest bias. Therefore, the out-of-sample prediction error $ERR_{oos}$ presents a more pronounced U-shape than $ERR_{is}$: due to the higher bias, the low-$k$ k-NN estimators perform worse out-of-sample than in-sample.

\paragraph{The effect of sampling noise.} This difference between in- and out-of-sample prediction becomes even more salient when we vary the outcome-noise level $\sigma$ in \cref{fig:sigma-effect}. (Note that when $\sigma$ changes, bias remains constant and only the variance term changes; see \cref{eq:bias,eq:var}.) In the absence of outcome noise ($\sigma=0$), k-NN estimators with higher complexity are \textit{always better} for in-sample prediction. This is \textit{not true}
 for out-of-sample prediction: at $\sigma=0$, the prediction error is equal to the bias term, which by \cref{fig:bias-var} itself has a U-shape in $k$ -- thus, k-NNs with intermediate level of complexity perform best at prediction \textit{even in the absence of outcome noise}. With $\sigma>0$, we observe a bias-variance tradeoff for in-sample prediction, while bias and variance lead to the \textit{the same out-of-sample ranking} of estimators for $1\leq k\leq 10$ (i.e. there is no tradeoff between bias and variance in this interval of $k$). In \cref{app:bias-decomp}, we also show that the terms $\text{NeighborMatchingBias}_k(x_0)$ and $\text{AveragingBias}_k(x_0)$ indeed behave as expected.

\subsection{Conclusion: The bias-variance tradeoff does not necessarily hold out-of-sample as it does in-sample}
In this section, we discovered that (and why) moving from in-sample prediction to out-of-sample prediction can have surprisingly stark effects on the classical textbook intuition relating model complexity to bias and variance. In particular, we demonstrated that while the bias-variance tradeoff intuition \textit{``Variance increases with model complexity, bias decreases with model complexity"} applies when considering the in-sample setting, it does not necessarily hold when considering out-of-sample prediction: we showed that there exist regimes where \textit{both} bias and variance decrease when model complexity is \textit{decreased}.\footnote{{Note that this is related to yet different from \citet{hastie2022surprises}, who show for linear regression that both (out-of-sample) bias and variance can decrease as model complexity \textit{increases} in the \textit{overparameterized} regime.}}

That is, we showed that classical statistical intuitions regarding the bias-variance tradeoff can break already in the (standard) underparametrized regime for extremely simple estimators and data-generating processes -- as a consequence of simply moving from in-sample prediction to out-of-sample prediction! As we show in the remainder of this note, this observation is crucial to understanding \textit{why} recently observed phenomena like double descent and benign overfitting contradict textbook wisdoms: overparameterization and interpolation are not responsible for breaking the bias-variance tradeoff intuitions on their own. Only in combination with a move in interest from fixed to random designs do the surprising modern phenomena arise.

\section{Reconciling double descent with textbook intuitions about the bias-variance tradeoff}\label{sec:dd}
\squish{The double descent phenomenon \cite{belkin2019reconciling} received considerable attention recently as its existence \textit{appears to} contradict the classical bias-variance tradeoff. In particular, \citet{belkin2019reconciling} highlighted that when plotting \textit{generalization error} against the total number of model parameters $p$, one observes a U-shaped error curve while $p\leq n$ (where error first decreases and then increases). Once one lets $p>n$, however, error experiences a \textit{second} descent in the so-called interpolation regime where the training data can be fit perfectly -- resulting in a double descent shape.}

In \citet{curth2023u}, we demonstrated that in the non-deep examples of double descent in \citet{belkin2019reconciling} (using trees, boosting and linear regressions), this non-monotonic behavior in error is directly caused by \textit{a change} in how model parameters are added at $p=n$. Further, we showed that once a measure for the \textit{effective parameters} \citep[Ch. 3.5]{hastie1990GAM} used by a model is placed on the x-axis, the double descent curves fold back into U-shaped curves, as adding additional raw model parameters in the interpolation regime leads to a decrease in test-time \textit{effective} parameters. In doing so, we already presented two resolutions to the ostensible tension between double descent and textbook intuitions regarding the bias-variance tradeoff. 

Here, we briefly wish to elaborate on a \textit{third} resolution -- more closely related to the topic of this note -- that we only alluded to in \citet[Appendix C.2]{curth2023u}. In particular, we wish to highlight that while \citet{belkin2019reconciling} argued that the historical absence of double descent curves in the statistics literature is likely due to a lack of the use of (unregularized) overparameterized models, there is a second reason we consider at least as important: as highlighted above, the statistics literature which developed many of the intuitions around the U-shaped curve historically considered mainly fixed-design settings, while double descent curves have been shown exclusively in random design settings. 

\begin{figure}[t]
    \centering
\includegraphics[width=0.8\linewidth]{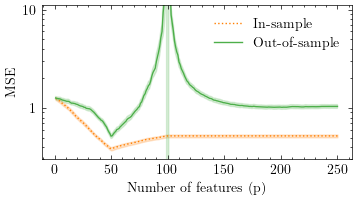}
    \caption{\textbf{A double descent shape appears only in out-of-sample prediction error, not in in-sample prediction error.} \small The behavior of in- and out-of-sample prediction error ($ERR_{is}$ and $ERR_{oos}$) as we vary the number of features $p$ included in a linear regression with $n=100$ training examples. \small In the underlying DGP, $\sigma=\frac{1}{2}$ and only the first $s=50$ features are used in $f^*$, all other $p-s$ features are irrelevant for prediction.  
    }
    \label{fig:dd}
\end{figure}

This is not a coincidence: it is easy to see that it is actually \textit{impossible} to observe a second descent in in-sample prediction error in the interpolation regime. That is, as the name suggests, any model in the interpolation regime issues predictions $\hat{f}(x_i)=y_i$ for training points $x_i$ regardless of how it is trained and parameterized (else it could not fit the training data perfectly). Thus, both in-sample bias ($=0$) and in-sample variance ($=\sigma^2$) are the same for any model in the interpolation regime, and as a consequence, so is in-sample prediction error. In fact, it is also easy to see that the in-sample prediction error of any interpolating model is $2\sigma^2$, regardless of the number of model parameters it uses. 

Using a linear regression experiment adapted from \citet{maddox2020rethinking}, we show this empirically in \cref{fig:dd}. Here, we fit a linear regression to an increasing number of features $d$ (resulting in $p=d$ parameters), where we use the ground truth DGP $f^*(x)=\frac{1}{\sqrt{s}}\sum^s_{k=l}x_s$ so that there are $d-s$ irrelevant features. We observe that, indeed, a double descent shape appears only in out-of-sample prediction (generalization) error; in-sample prediction error follows the familiar U-shape in the underparameterized regime and is constant in the overparameterized regime -- as is expected.

Strictly speaking, it is thus not really necessary to do any reconciling of double descent with the classical bias-variance tradeoff -- they appear in different settings and thus \textit{do not actually contradict each other}. Indeed, as we highlighted in the preceding section, even in the underparameterized regime the bias-variance tradeoff itself does not necessarily hold in the random design setting in the same way as it does in the fixed design setting. In this light, it should thus be less surprising that model behavior in the overparameterized regime differs too.

\section{\textit{Can} overfitting be benign? Understanding overfitting requires refining our vocabulary}\label{sec:overfitting}
Entangled in the literature on double descent is the so-called \textit{benign overfitting} effect \cite{bartlett2020benign} -- the observation that ML models can sometimes perform well despite fitting the training data \textit{perfectly}. This appears to stand in direct contradiction with textbook intuitions, which hold that ``a model with zero training error is overfit to the training
data and will typically generalize poorly''\citep[Ch. 7.2, p. 221]{hastie2009elements}.

In this section, we argue that understanding when and why ``overfitting'' can sometimes be benign requires more care and \textit{precision} in the vocabulary we use when discussing these phenomena. In fact, and somewhat surprisingly, \citet{bartlett2020benign} do not precisely define the term ``overfitting'' (and neither do \citet{hastie2009elements}). Instead, overfitting in the literature on ``benign overfitting'' appears to be used exchangeably with the term \textit{``interpolation''} -- which, in turn, refers to models that \textit{fit the training data perfectly}, i.e. attain zero training error $\hat{err}_{train}=\sum^n_{i=1} (y_i - \hat{f}(x_i))^2$. 

\squish{There is, however, a crucial difference between the term interpolation and the implied meaning of the term \textit{overfitting}. Indeed, the
Oxford dictionary defines ``overfitting (statistics)'' as: \textit{``The production of an analysis which corresponds too closely or exactly to a particular set of data, and may therefore fail to fit additional data or predict future observations reliably.''} Importantly, note that this definition implies that interpolation itself \textit{does not suffice} for overfitting -- the term \textit{over}fitting already implies that model performance \textit{suffers} due to interpolation. Instead, \citet{efron1997improvements}, for example, define the amount of overfitting as $ERR - \hat{err}_{train}$, the \textit{excess prediction error} (relative to the training error), also sometimes referred to as the degree of \textit{optimism} of the training error \citep{efron1986biased}.}

This makes immediately obvious that in the original sense of the term, it is \textit{impossible} for overfitting to be benign. In fact, ``benign overfitting'' is a tautology -- the term overfitting itself already implies that interpolation of the training data is malignant (reflected in $ERR >> \hat{err}_{train}$). Similar to e.g. \citet{muthukumar2020harmless}, who alternatively refer to the phenomenon as ``harmless interpolation'', we will therefore instead use the term \textit{benign interpolation} in the remainder.

Having cleared this initial semantic hurdle, we can now move to the statistical question at the heart of this section,  rephrasing it slightly: When can \textit{interpolation} be benign? That is, when are models that interpolate the training data overfit in the sense that their prediction performance is poor, and when can interpolating models perform well in terms of prediction? Intuitively, models that can interpolate any set of training data points have the capacity to fit pure noise, thus one may expect interpolation to always lead to overfitting. Below, we highlight that -- as before -- the answer to this question is strongly dependent on whether one is interested in the fixed or random design setting. 

\subsection{Can interpolation be benign in fixed design settings? (A: No!)}
If the (fixed design) in-sample prediction error is of interest, the answer is simple: interpolation cannot be benign. Indeed, by the bias-variance decomposition we trivially know that $ERR_{is}=Bias^2(\hat{f})+Var(\hat{f})+\sigma^2$. Because any interpolating model -- for example the 1-NN estimator -- predicts $\hat{f}(x_i)=y_i$, we have that $Bias(\hat{f})=f^*(x_i)-\mathbb{E}(\hat{f}(x_i)=f^*(x_i)-f^*(x_i) = 0$ and $Var(\hat{f})=\sigma^2$. Thus, the prediction error is dominated by the variance in outcome generation. In this case, the only time an interpolating solution is not overfit is if there is no noise in outcomes. Thus, overfitting in fixed design settings is caused by \textit{variance due to outcome noise alone}. 

\subsection{Can interpolation be benign in random design settings? (A: Yes, sometimes!)}
If you have made it this far in this note, you may not be surprised to discover that the switch to the random design setting changes also the answer to this question. Indeed, interpolating models can be either \textit{more or less} overfitted in the out-of-sample setting! 

\textbf{Why can overfitting behavior be \textit{worse} in the out-of-sample setting?} In the fixed design setting, interpolating models trivially have no bias. However, as we showed in \cref{sec:bias-bias-var}, models that interpolate the training data will generally have non-zero bias for a new test input $x_0$. Thus, interpolating models incur \textit{both bias and variance terms} in the out-of-sample setting. This is most easily seen by revisiting \cref{fig:bias-var,fig:sigma-effect}: the 1-NN estimator interpolates the training data but incurs \textit{no} $ERR_{is}$ in the absence of noise $\sigma$. It does, however, incur $ERR_{oos}$ even when there is no noise in outcomes because of non-zero bias. Thus, in this example for $\sigma=0$, $ERR_{is}=\hat{err}_{train}=0$ while $ERR_{oos}>>\hat{err}_{train}$. That is, the model is not overfit if a fixed design setting is of interest but it is overfit if a random design setting is of interest because -- unlike the fixed design setting where overfitting is due to variance alone -- in random design settings overfitting can be a consequence of \textit{both} the bias and the variance term! 

\textbf{Why can overfitting behavior be \textit{less pronounced} in the out-of-sample setting?} It is of course the opposite case that has received popular attention recently: \textit{some} interpolating models generalize well despite their ability to fit the training data perfectly \cite{zhang2021understanding, belkin2018overfitting, belkin2018understand, bartlett2020benign}. Intuitively, this is because some ML models can behave \textit{very differently} around new test points compared to inputs observed during training. This behavior is very different from e.g. classical k-NN estimators, which always use exactly k neighbors for prediction regardless of whether point was observed during training or not. In \citet{curth2024random}, we show that this is different for e.g. interpolating random forests, which act like 1-NN estimators on the training data, but can act like k-NN estimators with $k>1$ on previously unobserved inputs. \citet{wyner2017explaining} call this behavior \textit{spiked-smooth}, which provides a good metaphor: benignly interpolating models have the capacity to create sharp regions around training examples (`spike') where one may wish to retain precise knowledge of the known label but are much \textit{smoother} in regions of the input space where no information has been observed at training time and the problem is hence underdetermined. 

In recent work, we demonstrated that such a difference in behavior at train- and test-time is indeed quantifiable \textit{without access to test-time labels} by measuring the effective parameters \citep[Ch. 3.5]{hastie1990GAM} a model uses when issuing predictions. While such complexity measures were originally developed in the fixed design context and are therefore usually only computed for train-time predictions, we showed in \citet{curth2023u} that they can be adapted to the random design context and be computed for train and test-examples separately. We then showed that this can be used to predict when interpolation is benign in linear regression \cite{curth2023u}, random forests \cite{curth2024random} and even neural networks \cite{jeffares2024closer}, as in all cases benignly interpolating models use substantially less effective parameters when issuing predictions for new test examples than for train-time predictions. (To intuitively see why this may improve generalization performance, recall from \cref{sec:bias-var} that lowering model complexity can lead to a reduction in \textit{both bias and variance} for new test inputs.)

\section{Conclusion}
{In this note, we highlighted that one seldomly discussed yet immensely important factor in the emergence of modern  (apparently counterintuitive) ML phenomena is that model performance today is evaluated in terms of \textit{generalization to new inputs}, while the classical statistics literature, in which many of the intuitions regarding bias-variance tradeoffs and overfitting were developed, often considered \textit{in-sample prediction error} (where only noisy outcomes are resampled but input points are the same as during training). We showed that the move from fixed to random designs changes the classical bias-variance tradeoff even in the underparameterized regime for simple k-NN estimators, highlighting that behaviors that would appear to contradict classical statistical textbook intuitions can arise even in the absence of high-dimensional data, modern ML estimators and overparameterization -- factors that are usually held responsible for counterintuitive ML phenomena. We then demonstrated that this is another reason for the historical lack of observations of phenomena like double descent and benign overfitting: when fixed design prediction is of interest -- as was the case historically in statistics  --, it is impossible to observe such behavior.}

\paragraph{Implications. } Returning to the opening question ``Should we no longer be concerned about overfitting and bias-variance tradeoffs?'' the answer is thus ``It depends!''. It depends on the setting that is of interest in practice, and \textit{how} the used method interpolates the data (if it does). If generalization to new inputs is of interest -- as is the predominant setting in the modern ML literature -- and if models interpolate the training data in ways that are likely to be benign because predictions are smoother on test- than on training points -- as appears to be the case for e.g. neural networks and random forests -- then worrying about restricting the model's ability to perfectly fit the training data may no longer be necessary. 

If, however, training inputs are likely to reoccur at test-time then overfitting \textit{should be} a concern. This would be the case not only in classical fixed design settings where input points are somehow \textit{designed}, but also when observed inputs are \textit{coarser} than the latent variables that determine outcomes in the true underlying DGP -- e.g. when continuous characteristics are dichotomized during recording so that individuals with different underlying characteristics get mapped to the same input $x$. Another context in which training inputs can reoccur at (a different type of) test-time is in causal inference, where ML is sometimes used to impute nuisance functions that are then further processed in downstream analysis steps and therein evaluated at the training points \cite{van2011targeted, chernozhukov2018double}. There, sample-splitting is regularly used to forgo potential bias due to overfitting -- a practice which should indeed continue despite observations of benign interpolation in out-of-sample context. 

Finally, the contents of this note showcase that introductory textbooks and courses on statistical learning could potentially benefit from a makeover of their sections on the bias-variance tradeoff and overfitting, in particular, by being \textit{more precise about the settings} in which different intuitions are likely to apply (and why). Beyond questions relating to bias and variance, it would also be interesting and important to investigate whether the move from fixed to random designs affects any further fundamental statistical intuitions taught to statistics students around the world. 

\section*{Acknowledgements}
I would like to thank Alan Jeffares for countless thought-provoking discussions on the topic and for helpful comments on earlier versions of this note.

\appendix
\section*{Appendix}
\section{Bibliographical notes}
The primary purpose of this note is to be \textit{pedagogical}; it is hence written with only limited references to tangentially related work. Below, we briefly expand on some work that is related yet different from the topic of this note. 

The investigation contained in this note relates to and was partially inspired by \citet{rosset2019fixed}, who show that excess bias and variance terms appear in the bias-variance decomposition of the expected prediction error when comparing random to fixed designs and provide precise characterisations of these terms for the linear regression setting. (They do not, however, discuss implications for the bias-variance tradeoff and behavior of error as a function of model complexity or interpolation as we do here). 

With a similar goal of disentangling methodological complexity of modern ML methods from modern ML phenomena, \citet{belkin2018understand} show that benign interpolation is not unique to modern deep learning methods -- they show theoretically and empirically that benign interpolation appears in kernel methods too. Similarly, \citet{belkin2018overfitting} study generalization properties of a specific interpolating weighted nearest neighbor rule. \citet{belkin2019reconciling} showed that double descent occurs not only in deep neural networks were they were first observed \citep{bos1996dynamics}, but also in other more classical ML methods like linear regression.  Refer to \citet{loog2020brief} for a brief historical note on observations of double descent prior to \citet{belkin2019reconciling}.

\citet{neal2018modern} and \citet{neal2019bias} also note that textbooks may require an update regarding the bias-variance tradeoff as they empirically discover a lack of such tradeoff in deep neural networks, but do not link this to differences between in- and out-of-sample prediction. Instead, they highlight that in neural network training there are additional sources of variance beyond sampling noise in outcomes. \citet{adlam2020understanding} also revisit the bias-variance decomposition to better understand double descent. They provide a more fine-grained decomposition of the prediction error that takes into account all sources of randomness in modern ML, and show that this helps to understand deep double descent. (For the simple k-NN and linear regression methods that we consider here, this is not necessary as there are no such additional sources of randomness).

\citet{mallinar2022benign} provide a taxonomy and theoretical analysis of different classes of interpolating models, distinguishing between ``benign'', ``tempered'' and ``catastrophic'' behavior. There is a rich theoretical literature determining precise conditions when interpolation can be benign (for generalization error) in linear regression, see e.g. \citet{bartlett2020benign, muthukumar2020harmless, hastie2022surprises}.

 \begin{figure*}[!t]
	\centering
	\subfigure[Nonlinear DGP]{\centering\label{fig:nonlinear-biases}\includegraphics[width=0.42\textwidth]{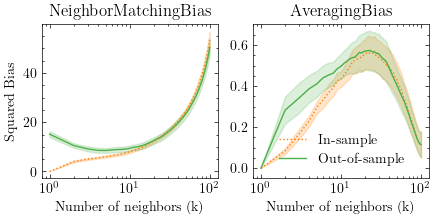}}
    \subfigure[Linear DGP]{\centering\label{fig:linear-bias}\includegraphics[width=0.42\textwidth]{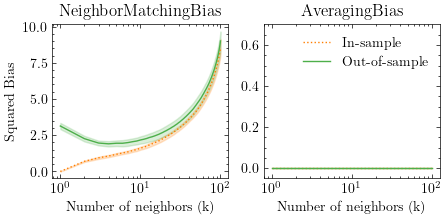}}
	\caption{\textbf{The bias due to lack of a perfect close neighbor match dominates the bias term out-of-sample. } \small The behavior of the Squared NeighborMatchingBias and Squared AveragingBias by k for kNN estimators, in-sample (orange) and out-of-sample (green) for a nonlinear (left) and linear DGP (right) }\label{fig:biases}   
\end{figure*}
\section{Empirically decomposing the bias term}\label{app:bias-decomp}

Here, we empirically investigate whether the terms $\text{NeighborMatchingBias}_k(x_0)$ and $\text{AveragingBias}_k(x_0)$ indeed behave as expected. As we discussed in \cref{sec:bias-bias-var}, the degree of nonlinearity of $f^*$ will play a role in this. Therefore, in addition to \cref{eq:marsmult},
we borrow a linear DGP from \citet{hastie2017extended}, 
\begin{equation}\label{eq:lin}
\textstyle f^*(x)=\sum^s_{l=1} x_l
\end{equation}
\squish{where only the first $s$ dimensions of $x \in \mathbb{R}^d$ (here: $s=5$, $d=10$) enter the regression specification and $x \sim \mathcal{N}(0, \Sigma)$, where the feature covariance matrix has $\Sigma_{ij}=0.35^{|i-j|}$.}

In \cref{fig:biases}, we observe that it is indeed the neighbor matching bias component that distinguishes in- and out-of-sample prediction: at $k=1$, decreasing the model complexity by adding an additional neighbor worsens the in-sample NeighborMatchingBias, but improves the out-of-sample bias. As expected, AveragingBias does not appear when the DGP is linear, but increases in $k$ in the nonlinear case as we add neighbors relative to the 1-NN estimator. (Note that it finally appears to decrease again once $k$ is very large; this may be due to an interplay of the DGP with the uniform distribution of inputs). Overall, for the DGPs considered here, it appears that it is the NeighborMatchingBias that dominates the total bias term (by orders of magnitude).



\bibliographystyle{icml2024}
\bibliography{example_paper}

\end{document}